%% file: emnlp2022.tex
\pdfoutput=1

\documentclass[11pt]{article}

\usepackage{header}

\usepackage{EMNLP2022}

\usepackage{times}
\usepackage{latexsym}

\usepackage[T1]{fontenc}

\usepackage[utf8]{inputenc}

\usepackage{microtype}

\usepackage{inconsolata}

\usepackage{booktabs}
\usepackage{makecell}
\usepackage{graphicx}
\usepackage{xcolor}
\usepackage{soul}
\usepackage{color}
\usepackage{url}
\usepackage{xspace}
\usepackage{parskip}
\usepackage{caption}
\usepackage{subcaption}
\usepackage{svg}
\usepackage{amsmath}
\usepackage{fdsymbol}
\usepackage{cleveref}

\newcommand{\rqtwo}{Q2\xspace}
\newcommand{\rqthree}{Q3\xspace}

\definecolor{darkpastelgreen}{rgb}{0.01, 0.75, 0.24}
\definecolor{seagreen}{rgb}{0.18, 0.55, 0.34}
\definecolor{mountainmeadow}{rgb}{0.19, 0.73, 0.56}
\definecolor{forestgreen}{rgb}{0.13, 0.55, 0.13}
\definecolor{fireenginered}{rgb}{0.81, 0.09, 0.13}
\definecolor{ballblue}{rgb}{0.13, 0.67, 0.8}
\definecolor{bleudefrance}{rgb}{0.19, 0.55, 0.91}

\definecolor{prompt}{HTML}{FCCDE5}
\definecolor{examples}{HTML}{D7F3E7}
\definecolor{tobeparaphrased}{HTML}{FFFFB3}
\definecolor{generated}{HTML}{B9DEFF}
\definecolor{testA}{HTML}{FFC2BA}
\definecolor{testB}{HTML}{CFCFCF}

\DeclareRobustCommand{\prompt}[1]{\sethlcolor{prompt}{\textbf{\hl{~#1~}}}}
\DeclareRobustCommand{\examples}[1]{\sethlcolor{examples}{\textbf{\hl{~#1~}}}}
\DeclareRobustCommand{\tobeparaphrased}[1]{\sethlcolor{tobeparaphrased}{\textbf{\hl{~#1~}}}}
\DeclareRobustCommand{\generated}[1]{\sethlcolor{generated}{\textbf{\hl{~#1~}}}}

\title{How Large Language Models are Transforming\\ Machine-Paraphrased Plagiarism}

\author{Jan Philip Wahle\textsuperscript{\includegraphics[scale=0.01]{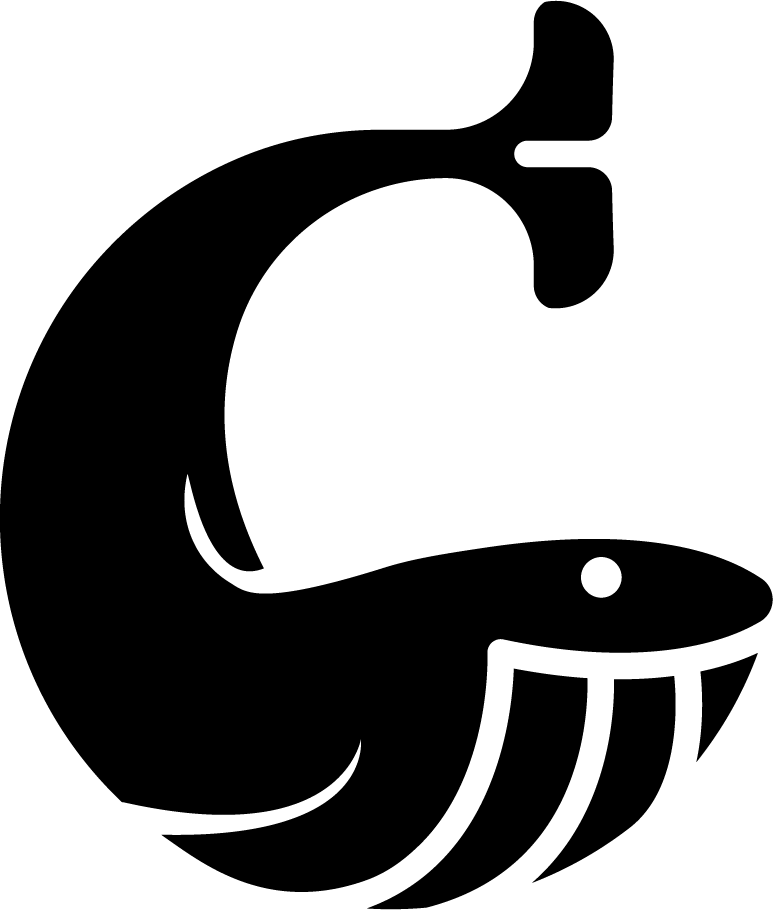}}\textsuperscript{*}, Terry Ruas\textsuperscript{*}, Frederic Kirstein$^{\spadesuit}$\textsuperscript{*}, Bela Gipp\textsuperscript{*} \\
\textsuperscript{*}Georg-August-Universität Göttingen, Germany\\
$^{\spadesuit}$Mercedes-Benz Group AG, Germany \\
\textsuperscript{\includegraphics[scale=0.01]{wahle-logo}}\texttt{wahle@gipplab.org}\\}

\begin{document}

\maketitle

\thispagestyle{firststyle}

\begin{abstract}
The recent success of large language models for text generation poses a severe threat to academic integrity, as plagiarists can generate realistic paraphrases indistinguishable from original work.
However, the role of large autoregressive transformers in generating machine-paraphrased plagiarism and their detection is still developing in the literature.
This work explores T5 and GPT-3 for machine-paraphrase generation on scientific articles from arXiv, student theses, and Wikipedia.
We evaluate the detection performance of six automated solutions and one commercial plagiarism detection software and perform a human study with 105 participants regarding their detection performance and the quality of generated examples.
Our results suggest that large models can rewrite text humans have difficulty identifying as machine-paraphrased (53\% mean acc.).
Human experts rate the quality of paraphrases generated by GPT-3 as high as original texts (clarity 4.0/5, fluency 4.2/5, coherence 3.8/5).
The best-performing detection model (GPT-3) achieves a 66\% F1-score in detecting paraphrases.
We make our code, data, and findings publicly available for research purposes.\footnote{\url{https://github.com/jpwahle/emnlp22-transforming}}
\end{abstract}

\section{Introduction}

\input{tables/showcase_main}
\textit{Paraphrases} are texts that convey the same meaning while using different words or sentence structures \cite{bhagat-hovy-2013-squibs}.
Paraphrasing plays an important role in related language understanding problems (e.g., question answering \cite{mccann2018natural}, summarization \cite{rush-etal-2015-neural}), but it can also be misused for academic plagiarism.
Academic plagiarism is serious misconduct as its perpetrators can unjustly advance their careers, obtain research funding that could be better spent, and make science less reliable if their misbehavior remains undetected \cite{meuschke_norman_2021_4913345}.

Paraphrasing tools can be used to generate convincing plagiarized texts with minimum effort.
Most of these tools (e.g., SpinBot\footnote{\url{https://spinbot.com/}}, SpinnerChief\footnote{\url{https://spinnerchief.com/}}) use relatively rudimentary heuristics, such as word replacements with synonyms, and they already deceive plagiarism detection software \cite{WahleRFM22}.
However, these tools scratch the surface of the possibilities compared to what large neural language models can achieve in producing convincing high-quality paraphrases \cite{zhou-bhat-2021-paraphrase}.
Notably, large autoregressive language models with billions of parameters, such as GPT-3 \cite{NEURIPS2020_1457c0d6}, make paraphrase plagiarism effortless yet exceedingly difficult to spot.

So far, large language models have found little application in plagiarism detection.
As language models are already easily accessible for applications such as software development\footnote{\url{https://copilot.github.com/}} or accounting\footnote{\url{https://openai.com/blog/openai-api/}}, using language models for machine-paraphrasing will become as easy as a click of a button soon.
Therefore, the number of machine-plagiarized texts will increase dramatically in the upcoming years.
To counteract this problem, we need robust solutions before models are widely misused.

In this study, we generate machine-paraphrased text with GPT-3 and T5 \cite{RaffelSRL20} to compose a dataset for testing against automatically generated paraphrasing.
We test different configurations of model size, training schemes, and selection criteria for generating paraphrases.
To understand how humans perceive machine-paraphrased text, we also performed an extensive study with 105 participants assessing their detection performance and quality-of-text assessments against existing automated detection methods.
We show that while humans can spot paraphrasing of online tools and smaller autoencoding models, large autoregressive models prove to be a more complex challenge as they can generate human-like text containing the same key ideas and messages from their original counterparts (see \Cref{table:showcase} for an example).
Popular paid plagiarism detection software (e.g. PlagScan\footnote{\url{https://www.plagscan.com/en/}}, Turnitin\footnote{\url{https://turnitin.com}}) is already deceived by rudimentary paraphrasing methods and large language models make this task even more challenging.
We also test the models used for the generation, which show the highest performance in detecting machine-paraphrased plagiarism.

To summarize our contributions:
\begin{itemize}
    \item We present a dataset with machine-paraphrased text from T5 and GPT-3 based on original work from Wikipedia, arXiv, and student theses to train and evaluate machine-paraphrased plagiarism.
    \item We explore the human ability to detect paraphrase through three experiments, focusing on (1) the detection difficulty of paraphrasing methods, (2) the quality of examples, and (3) the accuracy of humans in distinguishing between paraphrased and original texts.
    \item We empirically test plagiarism detection software (i.e., PlagScan) against machine learning methods and neural language models (autoencoding and autoregressive) in detecting machine-paraphrased plagiarism.
    \item We show that paraphrases from GPT-3 provide the most realistic plagiarism cases that both humans and automated detection solutions fail to spot, while the model itself is the best-tested candidate for detecting paraphrases.
\end{itemize}

\section{Related Work}
 
\noindent \textbf{Plagiarism Detection:} Plagiarism describes the use of ideas, concepts, words, or structures without proper source acknowledgment \cite{meuschke_norman_2021_4913345}.
Plagiarism datasets are limited to the number of real plagiarism cases known.
With the recent success of artificial intelligence in natural language processing (NLP) applications, paraphrase generation and plagiarism detection methods increasingly rely on dense text representations and machine learning classifiers \cite{FoltynekMG19}.
Machine learning methods often fail to detect substantial paraphrasing from neural language models \cite{Wahle2021}.
In particular, large autoregressive language models (e.g., GPT-3) can generate paraphrased content almost indistinguishable from original work \cite{witteveen-andrews-2019-paraphrasing}.
However, these models are still insufficiently explored in the domain of plagiarism detection, even though their impact on the field is already being discussed \cite{Dehouche21}.\\

\noindent \textbf{Machine-Paraphrase Detection:} Machine-paraphrasing can be described as the automatic generation of text that is semantically close to its source and written in other words \cite{bhagat-hovy-2013-squibs}.
Machine-paraphrasing experiences a growing research interest from NLP for learning semantic representations and related applications \cite{rush-etal-2015-neural, mccann2018natural}.
However, paraphrasing can be used in plagiarism detection to deceive humans and thus needs detection solutions to prevent it \cite{FoltynekMG19}.

Lexical substitution is a common paraphrase mechanism used by plagiarists \cite{BarronCedenoVMR13}.
Many online paraphrasing tools also use synonym replacements and other lexical perturbations to paraphrase text automatically \cite{FoltynekDAR20}.
\cite{FoltynekRSM20} showed that machine-learning classifiers (e.g., Support Vector Machine) could easily detect paraphrasing from popular online paraphrasing tools such as SpinBot.
\cite{Wahle2021} proposed a benchmark with paraphrased examples from autoencoding models (e.g., BERT\cite{DevlinCLT19}, RoBERTa\cite{LiuOMG2019}), showing that neural language models can generate more challenging paraphrasing than traditional online tools (e.g., SpinnerChief, SpinBot).
In a follow-up study, \cite{WahleRFM22} evaluate neural language models (e.g., BERT) on paraphrased texts from SpinnerChief, another independent paid online paraphrasing tool.
Their main finding was that neural language models outperform machine learning techniques and can obtain super-human performance in all test cases.
\cite{FoltynekRSM20,Wahle2021,WahleRFM22} results show that synonym replacements are simple to detect with state-of-the-art neural language models.
However, none of these studies explore using large autoregressive models in their experiments.

So far, only a few studies have analyzed the impacts of plagiarism using autoregressive models.
Seq2Seq models were first used by \cite{PrakashHLD} with stacked residual LSTM networks to generate paraphrases.
\cite{witteveen-andrews-2019-paraphrasing} train GPT-2 to generate paraphrased versions of a source text and select paraphrased candidates with the highest similarity according to universal sentence encoder\cite{CerYKH2018} embeddings and low word overlap when compared to their original counterparts. 
\cite{BidermanR22} show that GPT-J \cite{gpt-j}, a smaller version of GPT-3 with six billion parameters, can plagiarize student programming assignments that are not detected by MOSS\footnote{\url{https://theory.stanford.edu/~aiken/moss/}}, a popular plagiarism detection tool.
The scaling of models allows for the generation of text indistinguishable from human writing \cite{NEURIPS2020_1457c0d6}.
In addition, the models' increase in size and consequentially their performance \cite{KaplanMHB21} have the potential to make the paraphrase detection task even more difficult.

\section{Methodology}

This study focuses on understanding how humans and machines perceive large autoregressive machine-generated paraphrase examples.
Therefore, we first generate machine-paraphrased text with different model sizes of GPT-3 and T5.
We then generate a dataset composed of 200,000 examples from arXiv (20,966), Wikipedia (39,241), and student graduation theses (5,226) using the best configuration of both models.

We investigate how humans and existing detection solutions perceive this newly automated form of plagiarism.
In our human experiments, we compare paraphrased texts generated in this study to existing data that use paid online paraphrasing tools and autoencoding language models to paraphrase their texts.
Finally, we evaluate commercial plagiarism detection software, machine-learning classifiers, and neural language model-based approaches to the machine-paraphrase detection task.

\subsection{Paraphrase Generation}

\noindent\textbf{Method:} We generate candidate versions of paragraphs using prompts and human paraphrases as examples in a few-shot style prediction (\Cref{table:showcase_fewshot}).
We provide the model with the maximum number of human paraphrased examples that fit its context window with a maximum of 2048 tokens total. For both models, we use their default configuration.

Paraphrasing models' goal is to mimic human paraphrases.
Instead of manually engineering suitable prompts for the task, we use AutoPrompt \cite{shin-etal-2020-autoprompt} to determine task instructions based on the model's gradients.
As suggested by the authors, we place the predict-token at the end of our prompt.
One example of a generated prompt was ``Rephrase the following sentence.''
As humans tend to shorten text when paraphrasing, we limit the maximum number of generated tokens concerning the original version to 90\%, which is the approximate ratio of human plagiarism fragments in \cite{BarronCedenoVMR13}.
\Cref{table:showcase_fewshot} provides an example of the model's input/output when generating paraphrases.

\input{tables/showcase_fewshot}

\noindent\textbf{Candidate Selection:} Paraphrases that are similar to their source are of limited value as they have repetitive patterns, while those with high linguistic diversity often make models more robust \cite{qian-etal-2019-exploring}.
The quality of paraphrases is typically evaluated using three dimensions of quality (i.e., clarity, coherence, and fluency), where high-quality paraphrases are those with high semantic similarity and high lexical and syntactic diversity \cite{mccarthy2009components,zhou-bhat-2021-paraphrase}.
We aim to choose high-quality examples semantically close to the original content without reusing the exact words and structures \cite{witteveen-andrews-2019-paraphrasing}.

In this paper, we choose generated candidates that maximize their semantic similarity against their original counterparts while minimizing their count-based similarity.
We select the Pareto-optimal candidate that minimizes ROUGE-L and BLEU (i.e., penalizing the exact usage of words compared to the original version) and maximizes BERTScore \cite{zhang2019bertscore} and BARTScore\footnote{We use the large model version for both metrics.} \cite{yuan2021bartscore} (i.e., encouraging a similar meaning compared to the original version).
\Cref{table:showcase_candidate_selection} provides an example for generated paraphrases and their scores.
While examples with high count-based similarity usually convey the same essential message (e.g., \textbf{Out 1} and \textbf{Out 2}), they also share a similar sentence structure and word usage.
Examples with high semantic similarity and lower count-based similarity (e.g., \textbf{Out 3}) state the same meaning but rephrase the sentence with novel structure and similar words describing the same idea.

\noindent\textbf{Dataset Creation:} To provide data for common sources of academic plagiarism (i.e., scientific articles), we paraphrase the original examples of the machine paraphrase corpus (MPC) \cite{WahleRFM22} which is mainly composed of publications on arXiv, Wikipedia, and student's graduation theses.
As human-authored examples, we sample equally from two of the most popular paraphrase datasets, i.e., P4P and PPDB 2.0 \cite{zhou-bhat-2021-paraphrase}.
The P4P database \cite{BarronCedenoVMR13} is composed of realistic plagiarism cases with the paraphrase phenomena they contain (e.g., morphology-based, syntax-based, lexicon-based), and the PPDB 2.0 database \cite{pavlick-etal-2015-ppdb} is a large-scale paraphrase corpus extracted with bilingual pivoting from which we extract the high-quality phrasal and lexical subsets.

\input{tables/showcase_candidate_selection}

\subsection{Human Evaluation}

Our human study aims to understand how participants perceive machine-paraphrased plagiarism compared to original work and human-paraphrased text.
We used Amazon's Mechanical Turk (AMT) service to obtain human assessments for paraphrased text classification.
Additionally, we asked experts that actively published in the plagiarism detection domain over the past five years.
To have adequate statistical power in our analyses \cite{CardHKJ20}, we included a total of 105 participants (see \Cref{sec:appendix_b_demography} for details on demographic information about participants).

In the first part of the human study (\rqtwo in \Cref{sec:experiments}), 50 participants are provided with a mutually exclusive choice of whether a text was machine-paraphrased or original and a text field to justify their reasoning.
In the second part (\rqthree in \Cref{sec:experiments}), 50 participants from AMT and five experts from the research community were provided with a mutually exclusive choice of 5 points on a Likert scale for each of the three parameters of clarity, fluency, and coherence.
For the first experiment, each participant evaluated five texts for five models resulting in 1,250 text evaluations.
For the second experiment, each participant evaluated ten texts for three parameters, totaling 1,340 text evaluations.

Following common best practices on AMT \cite{BerinskyHL12}, evaluators had to have over a 95\% acceptance rate, be in the United States, and have completed over 1,000 successful tasks.
We excluded evaluators' assessments if their explanations were directly copied text from the task (> 90\% text match), did not match their classification, or were short, vague, or otherwise non-interpretable. 
Across experiments, 138 assessments ($\approx$10\%) were rejected and not included in the experiments.

\section{Research Questions \& Experiments}
\label{sec:experiments}

\begin{figure}[ht]
    \centering
    \includegraphics[width=\columnwidth]{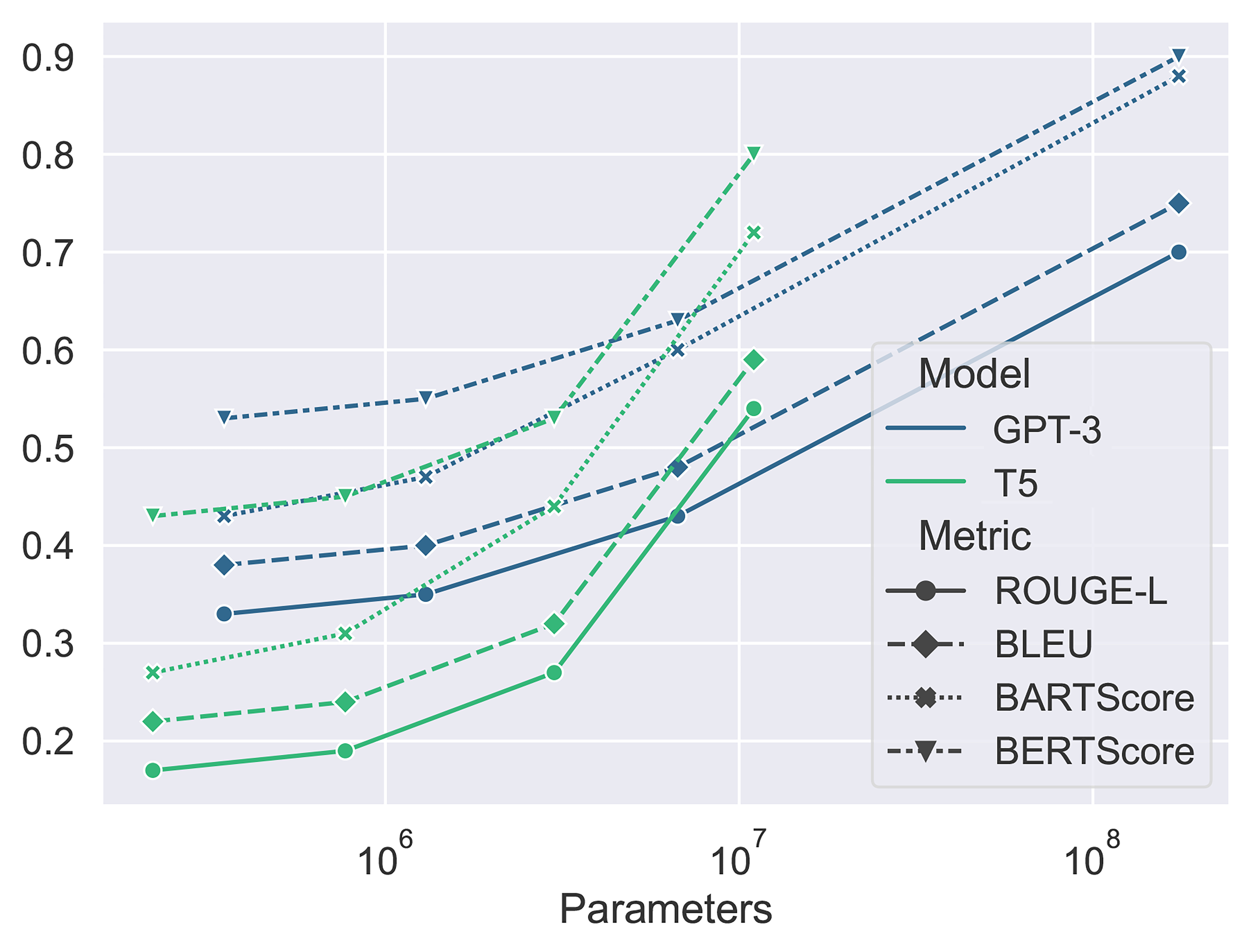}
    \caption{Paraphrasing similarity scores for a sample of the dataset with different model sizes of GPT-3 and T5.}
    \label{fig:exp_model_size}
\end{figure}

\noindent\textit{Q1: How does model size influence the quality of generated paraphrases?}\\[1.5pt]
A. We ask this question to underline the problem's urgency as recently released models have a large number of parameters.
\Cref{fig:exp_model_size} shows the influence of model size on the similarity scores of generated candidates against their original candidates on 500 random examples from the PPDB dataset.
We test the 220M, 770M, 3B, and 11B versions of T5 and the 350M, 1.3B, 6.7B, and 175B versions of GPT-3 (also known as Ada, Babbage, Curie, and Davinci in the OpenAI API\footnote{\url{https://openai.com/api/}} respectively).
With the increasing number of parameters, both models' semantic similarity scores (BERTScore, BARTScore) also rise. 
T5 shows the highest increase when extending the model from 3 billion parameters to 11 billion.
GPT-3 (175B) reaches its overall highest semantic similarity, generating sentences with similar meanings compared to the source.
Model's generated candidates also have higher count-based scores on average as they often repeat text from the source. As described before, we try to sample candidates with low word-count scores to avoid repetition of words.

We conclude that scaling models' size positively influences their performance at the task of paraphrasing, which agrees with previous research \cite{KaplanMHB21}.
While the limits and details of scaling models are still unknown, boosting their computing power will allow for more human-like texts to be produced.

\noindent\textit{Q2. Can humans identify whether a text is original, or machine-paraphrased?}\\[1.5pt]
A. This question is inspired by the \citet{Turing50} Test to differentiate machines from humans.
To answer this question, we asked participants to assess whether texts were machine-generated (see \Cref{sec:appendix_b_questions} for more details).
We compared original work to an online paraphrasing tool (SpinnerChief), two auto-encoding models (BERT, RoBERTa), and two large auto-regressive models (T5, GPT-3).
As examples, we sampled 30 machine-generated paragraphs for each model and their corresponding 30 original texts with an equal weighting between the three sources (Wikipedia, arXiv, and student theses).
We performed a Bonferroni-corrected two-sided T-Test to test for statistical significance compared to a control model.
As the control model, we chose SpinnerChief with its default paraphrasing frequency as it was the most difficult-to-detect online paraphrasing tool tested in \cite{WahleRFM22}.
Participants received individual text examples with three annotation options: ``machine-paraphrased'', ``original'', and ``I don't know''.
Participants were not shown aligned examples (i.e., an original and its paraphrased version) to avoid memorization effects.

\Cref{table:human_accuracy} shows the mean human accuracy (i.e., the ratio of correct assignments to non-neutral assignments per participant) in detecting machine-paraphrased text.
The results show that humans can adequately detect the control model with 82\% accuracy on average (where 50\% is a chance level performance).
In contrast, human accuracy at detecting paraphrases produced by autoencoding models was significantly lower, ranging from 61\% to 71\% over all participants.
Plagiarism cases generated by large autoregressive models were usually hardly above chance (53\% for GPT-3 and 56\% for T5).
For more information on the annotator agreement, please see \Cref{sec:appendix_b_agreement}.
Human abilities to detect machine-paraphrased text appear to decrease with increasing model size and are particularly challenging for autoregressive models as they can change sentence structure and word order instead of single word replacements.
Our findings on human detection against autoregressive models corroborate with recent results \cite{clark-etal-2021-thats}, challenging the common choice of humans as the gold standard.

\input{tables/human_eval_1}

\noindent\textit{Q3. How similar are machine-generated paraphrases to human-paraphrases?}\\[1.5pt]
A. We sampled 500 examples pairs (i.e., original, human-paraphrased) from the PPDB corpus and paraphrased half of the original versions with GPT-3 (175B) and the other half with T5 (11B).
As a proxy for similarity between originals, human-paraphrased, and machine-paraphrased examples, we calculated their similarity using BERTScore.
The average BERTScore between human-paraphrases and originals (76\%) is lower than between machine-generated paraphrases and originals (79\%).
The similarity between human-paraphrases and machine-generated paraphrases is highest (81\%).
This result suggest that machine-generated paraphrases are typically closer to the human paraphrases than to the original, which we assume is due to the model's objective to mimic human behavior, which are provided as generation examples.

\input{tables/human_eval_2}

\noindent\textit{Q4. How do humans assess the quality of machine-paraphrased plagiarism?}\\[1.5pt]
A. We asked human annotators to score generated paraphrases according to their clarity, fluency, and coherence \cite{zhou-bhat-2021-paraphrase} (see \Cref{sec:appendix_b_questions} for more details about the questions).
As quality assessments are challenging to evaluate, we increased the requirements for participants.
We asked the second group of 50 participants that required to have a higher education degree (bachelor's, master's, or Ph.D. degree).
We also asked additional five experts that have published at least two peer-reviews papers on plagiarism detection in the last five years.
Each participant annotated ten randomly drawn examples on a Likert scale from 1 to 5 regarding clarity, fluency, and coherence \cite{zhou-bhat-2021-paraphrase}.

\Cref{table:human_readability} shows the average rating for all 55 participants 
While original contents achieve the highest rating for all three dimensions, the largest version of GPT-3 achieves similar ratings.
SpinnerChief's quality of paraphrases is significantly lower.
BERT achieves convincing results as well, also because the frequency of word changes (15\%) for synonyms is lower than SpinnerChief's (50\%), and therefore generates examples closer to the original text.

Fluency was rated highest for all models, while clarity and coherence were the lowest.
We assume that as source sentences come from diverse scientific fields, they might already be difficult to understand; thus, paraphrasing can confuse readers when technical terms are used wrong. For more information on annotator agreement and the relation between experts and their educational degree, please see \Cref{sec:appendix_b_agreement}.

\input{tables/automated_eval}

\noindent\textit{Q5. How do existing detection methods identify paraphrased plagiarism?}\\[1.5pt]
A. To test the detection performance of automated plagiarism detection solutions, we evaluate five methods and compare them to random guesses and a human baseline.
We presume automated detection solutions can identify paraphrases better than humans as \cite{IppolitoDCE20a} showed that large language models are optimized to fool humans at the expense of introducing statistical anomalies which automated solutions can spot.
As a \textit{de-facto} solution for plagiarism detection, we test PlagScan, one of the best-performing systems, in a comprehensive test conducted by the European Network for Academic Integrity \cite{FoltynekDAR20}.
We test a combination of na\"{i}ve bayes classifier and word2vec \cite{MikolovCCD13}, and three autoencoding transformers: BERT \cite{DevlinCLT19}, RoBERTa\cite{LiuOMG2019}, and Longformer \cite{BeltagyPC20} which are the best performing models in machine-paraphrase detection of \cite{Wahle2021, WahleRFM22}.
Additionally, we evaluate the largest versions of T5 and GPT-3 using few-shot prediction.

As paraphrasing models, we choose SpinnerChief; the best performing paid online paraphrasing tool tested in \cite{WahleRFM22}.
Spinnerchief attempts to change every fourth word with a synonym.
We use BERT as an autoencoding baseline and set the masking probability to 15\% as in \cite{Wahle2021}.
As a large autoregressive model, we use GPT-3 175B, the best model for automated similarity metrics and deceiving humans.

\Cref{table:mainresult} shows the average F1-macro except for the human baseline, which shows accuracy.
For PlagScan, we assume positive examples when the text-match is greater than 50\%.
Looking at paraphrased plagiarism of SpinnerChief, humans reach between 79\% and 85\% accuracy on average.
PlagScan achieves results up to 7\% over the random baseline for Wikipedia articles but achieves close to random performance for student theses.
As in \cite{WahleRFM22}, we assume PlagScan indexes Wikipedia and arXiv but not student theses used in the MPC.
Neural approaches based on na\"{i}ve bayes reach between 58\% and 67\% F1-macro scores while autoencoding models achieve up to 67\% - 78\% (Longformer).
Large autoregressive models achieve peak scores of 85\% (T5 11B) and 87\% (GPT-3 175B) on SpinnerChief's paraphrases.

Results of detection models on BERT paraphrasing show similar patterns to SpinnerChief, as autoencoding models also replace masked words with synonyms.
While detection results are generally lower for humans and PlagScan, autoencoding models improve by a significant margin.
As pointed out in similar studies \cite{ZellersHRB19, Wahle2021}, models generating the paraphrased content are typically the best to detect it.
The similarity in the architecture of the autoencoding models allows BERT, RoBERTa, and Longformer for the largest performance increase over SpinnerChief.
Still, large autoregressive models achieve the best results in detecting machine-paraphrasing of BERT overall, with over 80\% F1-score for GPT-3.

When looking at paraphrasing of GPT-3, all models detect paraphrases significantly worse.
Humans, plagiarism detection software, and autoencoders can hardly achieve better results than random chance, which underlines how convincing paraphrased texts from large autoregressive models are.
T5 and GPT-3 can achieve low, but reasonable results between 60\% - 63\% (T5) and 64\% - 66\% (GPT-3) F1-macro.

While detection results on large autoregressive paraphrasing seem low, models were not explicitly trained on the task and are predicted based on previous fine-tuning on other data (upper part) or not fine-tuning (lower part).
We assume GPT-3 is the best detection solution because it generated the paraphrased texts.
Therefore, we see T5 as a baseline when autoregressive paraphrasing models are unknown.

In general, neural detection models reach their highest performance for Wikipedia articles which we assume is due to their pre-training data containing Wikipedia examples.
Student theses pose the most challenging scenario for both humans and neural approaches, as it contains challenging examples and is written by non-native English as a second language speakers.
Across experiments, PlagScan is not able to reliably identify machine-paraphrasing.
Large autoregressive models make it challenging for PlagScan to find text matches as phrasal and lexical substitutions can change the words with synonyms and the order of words.
The automatic detection results on paraphrasing of GPT-3 are alarming as many of the most used models fail to detect its paraphrases.
Even though the absolute results of GPT-3 and T5 are low, they can perform better than humans at the detection task.
Therefore, we assume that, similar to \cite{vahtola-etal-2021-coping}, there exist statistical abnormalities and patterns that automated solutions can leverage to increase their detection performance.

\section{Epilogue}

\noindent\textbf{Conclusion:} We generated machine-paraphrased plagiarism using large autoregressive models up to 175 billion parameters convincing paraphrased examples that deceived humans and plagiarism detection solutions.
We tested the human ability to detect machine-generated paraphrases of large models and compared their assessments to well-established online tools.
We evaluated one plagiarism detection software, one traditional machine-learning model, three autoencoding, and two large autoregressive models detecting machine-paraphrased examples.
Despite some limitations, our results suggest that large language models may increase the number of automated plagiarism cases through convincing paraphrasing of original work.

\noindent\textbf{Future Work:} This study is an initial step toward understanding how large language models can foster illicit activities in the scientific domain.
We plan to further examine the similarities and differences between human- and machine-generated paraphrases to understand whether humans have difficulties in detecting paraphrases in general.
When looking at participants' justifications for classifying machine-generated paraphrases, we plan to analyze common terms and highlights to find possible markers for classification decisions.
Over the scope of English, our approach could be applied to other languages and even generate paraphrases from one language to another using multilinugal models and data.
Finally, as academic plagiarism mainly relies on scientific articles, we want to extend our study to large scientific corpora with high variation across domains and venues \cite{lo-etal-2020-s2orc, Wahle2022c}.

\section*{Limitations}
Although our experiments explore how human and automated solutions struggle to identify machine-paraphrased examples from large language models, we did not detail the similarities and differences between human- and machine-generated paraphrases.
Comparing human paraphrases and machine paraphrases - qualitatively and automatically - would allow for a better understanding of what makes paraphrasing so challenging.
As the classification from our language models currently does not provide references or sources for their results, these models can only be used as a support tool to identify sentences and paragraphs for more detailed deliberation.
While our study has the above limitations, the focus of this study was to underline the urgency of the problem of machine-generated plagiarism to promote better detection solutions in the future.

\section*{Ethics Statement}
Plagiarism is illegal, unethical, and morally unacceptable in all countries \cite{kumar2013analysis}.
While the binary classification of machine-paraphrased examples in this study can indicate how automated detection solutions would point out potential plagiarism cases, a team of experts should make a final decision on such cases.
False-positive cases of wrongly accused researchers could ruin their careers forever. 
Therefore, all cases should be carefully evaluated before any final verdict.
As this study and related work show \cite{clark-etal-2021-thats}, humans are unreliable enough for paraphrase detection in the age of large neural language models.
The difficulty of machine-paraphrase identification makes legal decisions on plagiarism cases particularly complex.
We presume paraphrasing with language models will lead to more plagiarists getting unnoticed when using large models to generate their paraphrases.
One exciting approach to gain transparency would rely on reconstructing the model's potential inputs \cite{AAAI1714161,niu-etal-2019-bi} given the paraphrased version and classifying original candidates using a hybrid approach considering text-match and semantic features.
We adopted a binary classification in gender for our human evaluation, which we plan to improve in future work so it can be more inclusive.
Therefore, gender might not represent the natural diversity included in our dataset.

\bibliography{anthology,custom}
\bibliographystyle{acl_natbib}


\appendix

\section{Human Study}
\label{sec:appendix_b}

\subsection{Demographic Information of Participants}
\label{sec:appendix_b_demography}

Participants were given a choice to consent to providing additional anonymous information, including - but not limited to - gender, age, nationality, birth country, current country of residence, first language, and current education level\footnote{The complete list of attributes is available in our dataset.}.
Out of all 105 participants, 99 provided demographic information.
For all participants, we received their total number of completed tasks and the time taken to complete our questions.
The average time to rate ten examples was 8.07 ($\pm$ 6.82) minutes.
The average number of total successful tasks for participants was 1200 ($\pm$ 590).

The majority of tasks in this study were performed within 3 - 14 minutes (95\% of mass in the interval of [$\mu-2\sigma$, $\mu+2\sigma$]).
Three participants took significantly longer (23, 27, and 43 minutes), and their ratings were considered outliers on the distribution.

\begin{figure}[ht]
    \centering
    \includegraphics[width=\columnwidth]{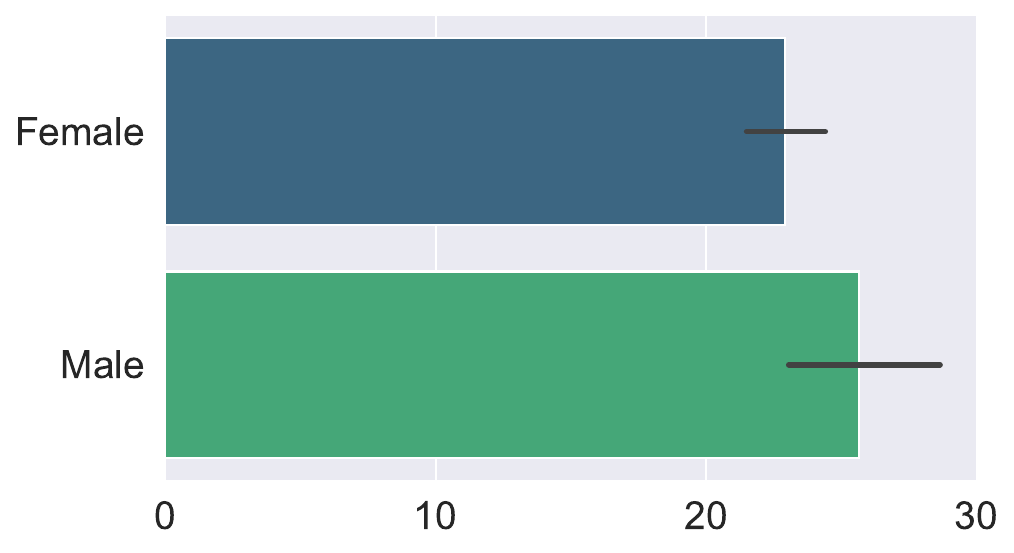}
    \caption{Distribution of age by gender of participants.}
    \label{fig:gender_age}
\end{figure}

\textbf{Age \& Gender:} Participants were 24 years old on average (18 - 41).
There was no significant difference in age between men and women with a two-sided T-Test (p=0.87).
\Cref{fig:gender_age} shows age distribution by gender.
The majority of participants were younger than 25 years old.

\textbf{Education \& First Language:}
Most participants from \textit{Q4} had a bachelor's degree (68\%).
The remainder had a master's degree (24\%) or Ph.D. degree (8\%).

Unsurprisingly, as all participants reside in the US, most of them (78\%) had English as their first language.
The remainder had Chinese, Spanish, Vietnamese, Russian, or Arabic as their first language.

\begin{figure}[ht]
    \centering
    \includegraphics[width=\columnwidth]{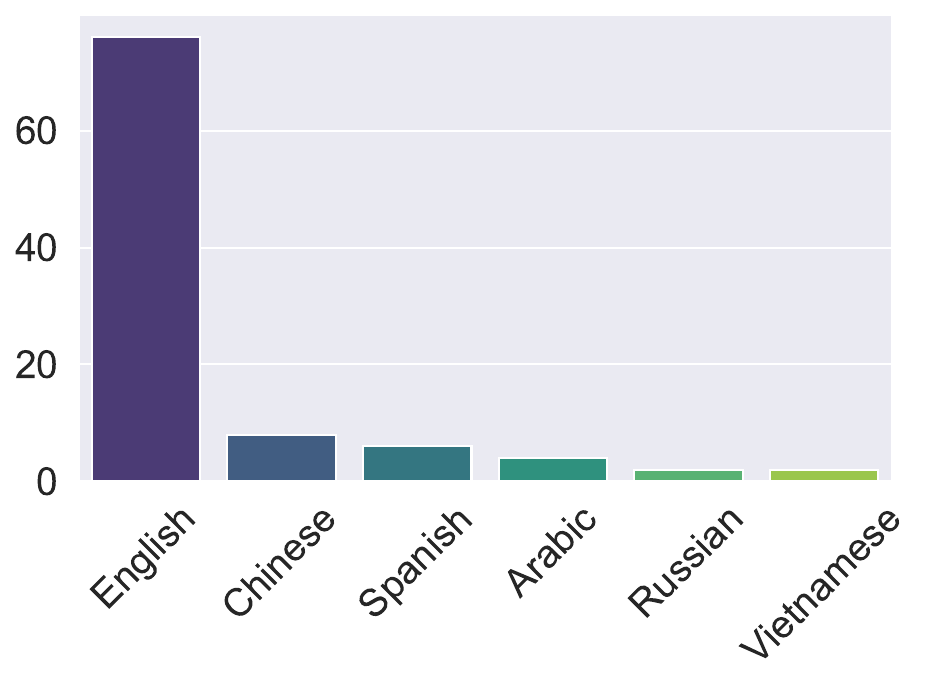}
    \caption{Distribution of first languages of participants.}
    \label{fig:language}
\end{figure}

\subsection{Agreement}
\label{sec:appendix_b_agreement}

The inter-annotator agreement according to Fleiss Kappa \cite{Fleiss73} of participants for \textit{Q2} was $\kappa = 0.84$.

The inter-annotator agreement of the five experts in \textit{Q4} was $\kappa = 0.66$ and for the remaining 50 participants in \textit{Q4} it was $\kappa = 0.79$.

The agreement between the expert group and the AMT group was $\kappa = 0.41$, showing that experts deviate strongly from average raters with a higher education degree.

When looking at participants with a Ph.D. and a bachelor's degree, assessments of paraphrasing quality deviated more $\kappa = 0.57$ than within the respective groups of participants with a Ph.D. degree $\kappa = 0.79$  and a master's degree  $\kappa = 0.77$.

\subsection{Details on Questions}
\label{sec:appendix_b_questions}

For the experiments in \textit{Q2}, participants were asked the following question: 

\textbf{Question:} Do you think the above example was machine-paraphrased (which means a machine rewrote some human-authored text) then choose ``machine-paraphrased''. If you think a human wrote the example, please choose ``original''. If you cannot assign the example to either category, please choose ``I don't know''.

For the experiments in \textit{Q4}, participants were given the following three instructions with the option to rate on a scale from one to five.

\textbf{Instruction 1:} The first question is about fluency, which refers to the ability to write grammatically correctly and clearly. Does it sound like a native speaker wrote it (high rating), or does it sound like someone who just learned English (low rating)?

\textbf{Instruction 2:} The second question is about clarity, which refers to the presentation of content and its explanation. Is the content easy to follow (high rating), or is it complicated and hard to understand (low rating)?

\textbf{Instruction 3:} The third question is about coherence, which refers to the consistency of content throughout the paragraph. Is the content following a common central idea (high rating), or is the text jumping from one (random) idea to another (low rating)?

\end{document}

%% file: tables/showcase_main.tex
\begin{table}[!htb]
\small
\resizebox{\columnwidth}{!}{
    \begin{tabular}{l}
    \toprule
    \textbf{Original Text} \\ 
    \midrule
    \makecell[l]{... \\
    On \textbf{\textcolor{fireenginered}{April 29, 2017}}, \textcolor{blue}{\textbf{Bill Gates}} partnered with\\
    Swiss tennis legend \textcolor{forestgreen}{\textbf{Roger Federer}} in playing the\\
    ``Match for \textbf{\textcolor{bleudefrance}{Africa}}'' 4, a noncompetitive tennis match\\
    at a sold-out Key Arena in Seattle.\\
    The event was in support of \textcolor{forestgreen}{\textbf{Roger Federer}}\\
    Foundation's charity efforts in \textbf{\textcolor{bleudefrance}{Africa}}.\\
    ...}\\
    \midrule
    \textbf{Paraphrased using GPT-3} \\
    \midrule
    \makecell[l]{... \\
    \textcolor{blue}{\textbf{Bill Gates}} teamed up with Swiss tennis player\\
    \textcolor{forestgreen}{\textbf{Roger Federer}} to play in the ``Match for \textbf{\textcolor{bleudefrance}{Africa}} 4'' on\\ \textbf{\textcolor{fireenginered}{April 29, 2017}}.\\
    The noncompetitive tennis match at a sold-out\\ 
    Key Arena in Seattle was in support of\\ \textcolor{forestgreen}{\textbf{Roger Federer}} Foundation's charity efforts in \textbf{\textcolor{bleudefrance}{Africa}}.\\
    ...}\\
    \bottomrule
    \end{tabular}
}
\caption{Example excerpt from a Wikipedia article and its paraphrased versions using GPT-3. Important keywords are highlighted in boldfont and color. Autoregressive paraphrasing with GPT-3 keeps the same message while generating text with the original structure. The original example used is \texttt{3747-ORIG-44.txt}.}
\label{table:showcase}
\end{table}

%% file: tables/showcase_fewshot.tex
\begin{table}[t]
\small
\resizebox{\columnwidth}{!}{
    \begin{tabular}{l}
    \toprule
    \textbf{Paraphrase Generation Example} \\ 
    \midrule
    \makecell[l]{
    {\textbf{\prompt{Rephrase the following paragraph while keeping its meaning:}}}\\\\
    Original: {\textbf{\examples{My day has been pretty good.}}}\\
    Paraphrased: {\textbf{\examples{Today was a good day.}}}\\\\
    $\cdots$\\\\
    Original: {\textbf{\examples{This paper analyses two paraphrasing methods.}}}\\
    Paraphrased: {\textbf{\examples{We analyze two methods in this study.}}}\\\\
    \\Original: {\textbf{\tobeparaphrased{This text was written by a machine.}}}\\
    Paraphrased: {\textbf{\generated{This sentence has been generated artificially.}}}\\
    }\\
    \bottomrule
    \end{tabular}
}
\caption{Example of generating paraphrased plagiarism with few-shot learning. As input the model receives a \textbf{\prompt{prompt}} and human paraphrase \textbf{\examples{example pairs}}. After inserting the \textbf{\tobeparaphrased{to-be-paraphrased sentence}}, the model then generates a \textbf{\generated{paraphrased version}} as the output. }
\label{table:showcase_fewshot}
\end{table}

%% file: tables/showcase_candidate_selection.tex
\begin{table*}[t!]
    \centering
    \resizebox{\textwidth}{!}{
        \begin{tabular}{lp{1.2\columnwidth}cccc} 
        \toprule							
         &\multicolumn{1}{r}{} & BERTSc. & BARTSc. & Rouge-L & BLEU   \\				
        \midrule
        \textbf{\small{In:}} & Later in his career, Gates has pursued many business and philanthropic endeavors. & - & - & - & -   \\
        \midrule
        \textbf{\small{Out 1:}} & Later, his time was allocated to business and philanthropic endeavors. & 0.79 & 0.74 & 0.55 & 0.63   \\
        \midrule
        \textbf{\small{Out 2:}} & Later in his career, Gates focused on business and charity. & 0.84 & 0.83 & 0.64 & 0.51    \\
        \midrule
        \textbf{\small{Out 3*:}} & \textbf{Gates focused on business and charitable efforts later in his career.} & \textbf{0.83} & \textbf{0.85} & \textbf{0.35} & \textbf{0.49} \\
        \midrule
        \end{tabular}
    }
   \caption{Candidate selection of machine-generated paraphrases with an example from \cite{witteveen-andrews-2019-paraphrasing}. We choose the Pareto-optimal example that maximizes semantic similarity (BERTScore, BARTScore) and minimizes word overlap (ROUGE-L, BLEU). $^*$Selected example in boldface.}
    \label{table:showcase_candidate_selection}
\end{table*}

%% file: tables/human_eval_1.tex
\begin{table*}[!ht]
    \centering
        \begin{tabular}{l c c c c}
        \toprule
         & \makecell{Mean accuracy} & \makecell{95\% Confidence \\Interval (low, hi)} & \makecell{$t$ compared to \\ control ($p$-value)} & \makecell{``I don\'t know" \\ assignments} \\ 
        \midrule
        SpinnerChief (Control)          & 82\% & 76\%-89\% & - & 2.8 \% \\ 
        \midrule
        BERT                            & 67\% & 63\%–71\% & 14.2 (1$e$-11) & 4.9\% \\ 
        RoBERTa                         & 65\% & 61\%–70\% & 18.1 (1$e$-29) & 5.5\% \\ 
        \midrule
        T5 11B                          & 56\% & 51\%–59\% & 16.6 (1$e$-16) & 7.1\% \\ 
        \textbf{GPT-3 175B}             & \textbf{53\%} & \textbf{49\%–55\%} & \textbf{19.2} (1$e$-34) & 7.2\% \\ 
        \bottomrule
        \end{tabular}
    \caption{Human accuracy in identifying whether parapgraphs of scientific papers from the arXiv subset are machine-paraphrased. Human performance ranges from 82\% on the control model to 53\% on GPT-3 175B. This table compares mean accuracy of with five paraphrasing models and shows the results of a two-sample T-Test between each model and the SpinnerChief control model according to \cite{WahleRFM22}. Lowest scores are in \textbf{boldface}.}
    \label{table:human_accuracy}
\end{table*}

%% file: tables/human_eval_2.tex
\begin{table}[ht!]
    \centering
        \resizebox{\columnwidth}{!}{
            \begin{tabular}{l c c c}
            \toprule
             & Clarity & Fluency & Coherence\\ 
            \midrule
            Original        & 
            3.98 ($\pm$0.78) & 4.21 ($\pm$0.81) & 3.81 ($\pm$0.92) \\ 
            SpinnerChief    & 2.52 ($\pm$1.15) & 2.94 ($\pm$1.19) & 2.83 ($\pm$1.23) \\ 
            BERT            & 3.45 ($\pm$.1.29) & 3.34 ($\pm$0.90) & \textbf{3.73} ($\pm$1.22) \\ 
            \textbf{GPT-3}  & \textbf{3.92} ($\pm$0.97) & \textbf{3.60} ($\pm$1.02) & 3.72 ($\pm$1.07)\\
            \bottomrule
            \end{tabular}
        }
    \caption{Average scores on a Likert-scale from 1 to 5 of machine-generated plagiarism on the Wikipedia test set. Each example is judged by 50 participants with a bachelor's, master's, or PhD degree and five experts in the plagiarism detection community. Standard deviation is shown in parenthesis. Highest scores are in \textbf{boldface}.}
    \label{table:human_readability}
\end{table}

%% file: tables/automated_eval.tex
\begin{table*}[!ht]
\centering
\resizebox{\textwidth}{!}{
    \begin{tabular}{llllllllll}
        \toprule
        & \multicolumn{3}{c}{SpinnerChief} & \multicolumn{3}{c}{BERT} & \multicolumn{3}{c}{GPT-3} \\
        \midrule
        Model & arXiv & Theses & Wiki & arXiv & Theses & Wiki & arXiv & Theses & Wiki \\
        \midrule
        Random & 51.72 & 53.23 & 49.21 & 51.90 & 50.24 & 48.28 & 50.61 & 50.30 & 49.77 \\
        Human Baseline$^\dagger$ & 83.25 & 79.32 & 84.96 & 68.93 & 63.41 & 69.08 & 55.74 & 50.60 & 52.82 \\
        PlagScan$^{\dagger\dagger}$ & 55.07 & 49.29 & 57.10 & 57.73 & 50.22 & 59.04 & 49.28 & 48.90 & 50.19 \\
        w2v + NB & 65.89 & 58.24 & 66.83 & 62.12 & 59.96 & 63.38 & 52.78 & 51.01 & 51.15 \\
        BERT & 64.59 & 63.59 & 57.45 & 80.83 & 74.74 & 83.21 & 52.44 & 50.89 & 52.59 \\
	    RoBERTa & 66.00 & 58.24 & 58.94 & 70.41 & 68.99 & 72.18 & 53.14 & 49.90 & 53.81 \\        
	    Longformer & 78.34 & 74.82 & 67.11 & 65.18 & 65.72 & 69.98 & 54.70 & 50.84 & 53.99 \\
        \midrule
        T5 11B & \textbf{82.92$^{**}$} & \textbf{83.45$^{**}$} & \textbf{79.92$^{**}$} & \textbf{84.66$^{**}$} & \textbf{78.09$^{**}$} & \textbf{82.37$^{**}$} & \textbf{59.80$^{**}$} & \textbf{61.42$^{**}$} & \textbf{62.72$^{**}$}\\
        GPT-3 175B & \textbf{83.20$^{**}$} & \textbf{82.11$^{**}$} & \textbf{79.68$^{**}$} & \textbf{87.21$^{**}$} & \textbf{81.02$^{**}$} & \textbf{84.48$^{**}$} & \textbf{66.52$^{**}$} & \textbf{64.38$^{**}$} & \textbf{65.79$^{**}$} \\
    \bottomrule
    \end{tabular}
}
\caption{F1-Macro scores of detection models for text paraphrased by SpinnerChief, BERT, and GPT-3. Numbers in \textbf{boldface} are the overall best result. $^{**}$Results are statistically significant using random and permutation tests \cite{dror-etal-2018-hitchhikers} with p < 0.05. \textsuperscript{\textdagger}Accuracy calculated  as in \Cref{table:human_accuracy}. \textsuperscript{\textdagger}\textsuperscript{\textdagger}F1-score when text-match is greter than 50\%.} 
\label{table:mainresult}
\end{table*}

%% file: emnlp2022.bbl
\begin{thebibliography}{44}
\expandafter\ifx\csname natexlab\endcsname\relax\def\natexlab#1{#1}\fi

\bibitem[{{Barr{\'o}n-Cede{\~n}o} et~al.(2013){Barr{\'o}n-Cede{\~n}o}, Vila,
  Mart{\'i}, and Rosso}]{BarronCedenoVMR13}
Alberto {Barr{\'o}n-Cede{\~n}o}, Marta Vila, M.~Mart{\'i}, and Paolo Rosso.
  2013.
\newblock \href {https://doi.org/10.1162/COLI_a_00153} {Plagiarism {{Meets
  Paraphrasing}}: {{Insights}} for the {{Next Generation}} in {{Automatic
  Plagiarism Detection}}}.
\newblock \emph{Computational Linguistics}, 39(4):917--947.

\bibitem[{Beltagy et~al.(2020)Beltagy, Peters, and Cohan}]{BeltagyPC20}
Iz~Beltagy, Matthew~E. Peters, and Arman Cohan. 2020.
\newblock \href {http://arxiv.org/abs/2004.05150} {Longformer: The
  long-document transformer}.

\bibitem[{Berinsky et~al.(2012)Berinsky, Huber, and Lenz}]{BerinskyHL12}
Adam~J. Berinsky, Gregory~A. Huber, and Gabriel~S. Lenz. 2012.
\newblock \href {https://doi.org/10.1093/pan/mpr057} {Evaluating online labor
  markets for experimental research: Amazon.com's mechanical turk}.
\newblock \emph{Political Analysis}, 20(3):351–368.

\bibitem[{Bhagat and Hovy(2013)}]{bhagat-hovy-2013-squibs}
Rahul Bhagat and Eduard Hovy. 2013.
\newblock \href {https://doi.org/10.1162/COLI_a_00166} {{S}quibs: What is a
  paraphrase?}
\newblock \emph{Computational Linguistics}, 39(3):463--472.

\bibitem[{Biderman and Raff(2022)}]{BidermanR22}
Stella Biderman and Edward Raff. 2022.
\newblock \href {http://arxiv.org/abs/2201.07406} {Neural {{Language Models}}
  are {{Effective Plagiarists}}}.
\newblock \emph{ArXiv220107406 Cs}.

\bibitem[{Brown et~al.(2020)Brown, Mann, Ryder, Subbiah, Kaplan, Dhariwal,
  Neelakantan, Shyam, Sastry, Askell, Agarwal, {Herbert-Voss}, Krueger,
  Henighan, Child, Ramesh, Ziegler, Wu, Winter, Hesse, Chen, Sigler, Litwin,
  Gray, Chess, Clark, Berner, McCandlish, Radford, Sutskever, and
  Amodei}]{NEURIPS2020_1457c0d6}
Tom Brown, Benjamin Mann, Nick Ryder, Melanie Subbiah, Jared~D Kaplan, Prafulla
  Dhariwal, Arvind Neelakantan, Pranav Shyam, Girish Sastry, Amanda Askell,
  Sandhini Agarwal, Ariel {Herbert-Voss}, Gretchen Krueger, Tom Henighan, Rewon
  Child, Aditya Ramesh, Daniel Ziegler, Jeffrey Wu, Clemens Winter, Chris
  Hesse, Mark Chen, Eric Sigler, Mateusz Litwin, Scott Gray, Benjamin Chess,
  Jack Clark, Christopher Berner, Sam McCandlish, Alec Radford, Ilya Sutskever,
  and Dario Amodei. 2020.
\newblock Language {{Models}} are {{Few-Shot Learners}}.
\newblock In \emph{Advances in {{Neural Information Processing Systems}}},
  volume~33, pages 1877--1901. {Curran Associates, Inc.}

\bibitem[{Card et~al.(2020)Card, Henderson, Khandelwal, Jia, Mahowald, and
  Jurafsky}]{CardHKJ20}
Dallas Card, Peter Henderson, Urvashi Khandelwal, Robin Jia, Kyle Mahowald, and
  Dan Jurafsky. 2020.
\newblock \href {https://doi.org/10.18653/v1/2020.emnlp-main.745} {With
  {{Little Power Comes Great Responsibility}}}.
\newblock In \emph{Proceedings of the 2020 {{Conference}} on {{Empirical
  Methods}} in {{Natural Language Processing}} ({{EMNLP}})}, pages 9263--9274,
  {Online}. {Association for Computational Linguistics}.

\bibitem[{Cer et~al.(2018)Cer, Yang, Kong, Hua, Limtiaco, John, Constant,
  Guajardo-Cespedes, Yuan, Tar, Sung, Strope, and Kurzweil}]{CerYKH2018}
Daniel Cer, Yinfei Yang, Sheng-yi Kong, Nan Hua, Nicole Limtiaco, Rhomni~St.
  John, Noah Constant, Mario Guajardo-Cespedes, Steve Yuan, Chris Tar,
  Yun-Hsuan Sung, Brian Strope, and Ray Kurzweil. 2018.
\newblock \href {https://doi.org/10.48550/ARXIV.1803.11175} {Universal sentence
  encoder}.

\bibitem[{Clark et~al.(2021)Clark, August, Serrano, Haduong, Gururangan, and
  Smith}]{clark-etal-2021-thats}
Elizabeth Clark, Tal August, Sofia Serrano, Nikita Haduong, Suchin Gururangan,
  and Noah~A. Smith. 2021.
\newblock \href {https://doi.org/10.18653/v1/2021.acl-long.565} {All that{'}s
  {`}human{'} is not gold: Evaluating human evaluation of generated text}.
\newblock In \emph{Proceedings of the 59th Annual Meeting of the Association
  for Computational Linguistics and the 11th International Joint Conference on
  Natural Language Processing (Volume 1: Long Papers)}, pages 7282--7296,
  Online. Association for Computational Linguistics.

\bibitem[{Dehouche(2021)}]{Dehouche21}
N~Dehouche. 2021.
\newblock \href {https://doi.org/10.3354/esep00195} {Plagiarism in the age of
  massive {{Generative Pre-trained Transformers}} ({{GPT-3}})}.
\newblock \emph{Ethics. Sci. Environ. Polit.}, 21:17--23.

\bibitem[{Devlin et~al.(2019)Devlin, Chang, Lee, and Toutanova}]{DevlinCLT19}
Jacob Devlin, Ming-Wei Chang, Kenton Lee, and Kristina Toutanova. 2019.
\newblock \href {http://arxiv.org/abs/1810.04805} {{{BERT}}: {{Pre-training}}
  of {{Deep Bidirectional Transformers}} for {{Language Understanding}}}.
\newblock \emph{ArXiv181004805 Cs}.

\bibitem[{Dror et~al.(2018)Dror, Baumer, Shlomov, and
  Reichart}]{dror-etal-2018-hitchhikers}
Rotem Dror, Gili Baumer, Segev Shlomov, and Roi Reichart. 2018.
\newblock \href {https://doi.org/10.18653/v1/P18-1128} {The hitchhiker{'}s
  guide to testing statistical significance in natural language processing}.
\newblock In \emph{Proceedings of the 56th Annual Meeting of the Association
  for Computational Linguistics (Volume 1: Long Papers)}, pages 1383--1392,
  Melbourne, Australia. Association for Computational Linguistics.

\bibitem[{Fleiss and Cohen(1973)}]{Fleiss73}
Joseph~L. Fleiss and Jacob Cohen. 1973.
\newblock \href {https://doi.org/10.1177/001316447303300309} {The equivalence
  of weighted kappa and the intraclass correlation coefficient as measures of
  reliability}.
\newblock \emph{Educational and Psychological Measurement}, 33(3):613--619.

\bibitem[{Folt{\'y}nek et~al.(2020{\natexlab{a}})Folt{\'y}nek, Dlabolov{\'a},
  {Anohina-Naumeca}, Raz{\i}, Kravjar, Kamzola, {Guerrero-Dib}, {\c C}elik, and
  {Weber-Wulff}}]{FoltynekDAR20}
Tom{\'a}{\v s} Folt{\'y}nek, Dita Dlabolov{\'a}, Alla {Anohina-Naumeca}, Salim
  Raz{\i}, J{\'u}lius Kravjar, Laima Kamzola, Jean {Guerrero-Dib},
  {\"O}zg{\"u}r {\c C}elik, and Debora {Weber-Wulff}. 2020{\natexlab{a}}.
\newblock \href {http://arxiv.org/abs/2002.04279} {Testing of {{Support Tools}}
  for {{Plagiarism Detection}}}.
\newblock \emph{ArXiv200204279 CsDL}.

\bibitem[{Folt{\'y}nek et~al.(2019)Folt{\'y}nek, Meuschke, and
  Gipp}]{FoltynekMG19}
Tom{\'a}{\v s} Folt{\'y}nek, Norman Meuschke, and Bela Gipp. 2019.
\newblock \href {https://doi.org/10/dcxc} {Academic {{Plagiarism Detection}}:
  {{A Systematic Literature Review}}}.
\newblock \emph{ACM Computing Surveys}, 52(6):112:1--112:42.

\bibitem[{Folt{\'y}nek et~al.(2020{\natexlab{b}})Folt{\'y}nek, Ruas, Scharpf,
  Meuschke, Schubotz, Grosky, and Gipp}]{FoltynekRSM20}
Tom{\'a}{\v s} Folt{\'y}nek, Terry Ruas, Philipp Scharpf, Norman Meuschke,
  Moritz Schubotz, William Grosky, and Bela Gipp. 2020{\natexlab{b}}.
\newblock \href {https://doi.org/10.1007/978-3-030-43687-2_68} {Detecting
  {{Machine-Obfuscated Plagiarism}}}.
\newblock In Anneli Sundqvist, Gerd Berget, Jan Nolin, and Kjell~Ivar
  Skjerdingstad, editors, \emph{Sustainable {{Digital Communities}}}, volume
  12051 LNCS, pages 816--827. {Springer International Publishing}, {Cham}.

\bibitem[{Ippolito et~al.(2020)Ippolito, Duckworth, {Callison-Burch}, and
  Eck}]{IppolitoDCE20a}
Daphne Ippolito, Daniel Duckworth, Chris {Callison-Burch}, and Douglas Eck.
  2020.
\newblock \href {https://doi.org/10.18653/v1/2020.acl-main.164} {Automatic
  {{Detection}} of {{Generated Text}} is {{Easiest}} when {{Humans}} are
  {{Fooled}}}.
\newblock In \emph{Proceedings of the 58th {{Annual Meeting}} of the
  {{Association}} for {{Computational Linguistics}}}, pages 1808--1822,
  {Online}. {Association for Computational Linguistics}.

\bibitem[{Kaplan et~al.(2020)Kaplan, McCandlish, Henighan, Brown, Chess, Child,
  Gray, Radford, Wu, and Amodei}]{KaplanMHB21}
Jared Kaplan, Sam McCandlish, Tom Henighan, Tom~B. Brown, Benjamin Chess, Rewon
  Child, Scott Gray, Alec Radford, Jeffrey Wu, and Dario Amodei. 2020.
\newblock \href {https://doi.org/10.48550/ARXIV.2001.08361} {Scaling laws for
  neural language models}.

\bibitem[{Kumar and Tripathi(2013)}]{kumar2013analysis}
Ranjeet Kumar and RC~Tripathi. 2013.
\newblock An analysis of automated detection techniques for textual similarity
  in research documents.
\newblock \emph{International Journal of Advanced Science and Technology},
  56(9):99--110.

\bibitem[{Liu et~al.(2019)Liu, Ott, Goyal, Du, Joshi, Chen, Levy, Lewis,
  Zettlemoyer, and Stoyanov}]{LiuOMG2019}
Yinhan Liu, Myle Ott, Naman Goyal, Jingfei Du, Mandar Joshi, Danqi Chen, Omer
  Levy, Mike Lewis, Luke Zettlemoyer, and Veselin Stoyanov. 2019.
\newblock \href {http://arxiv.org/abs/1907.11692} {Roberta: A robustly
  optimized bert pretraining approach}.
\newblock Cite arxiv:1907.11692.

\bibitem[{Lo et~al.(2020)Lo, Wang, Neumann, Kinney, and
  Weld}]{lo-etal-2020-s2orc}
Kyle Lo, Lucy~Lu Wang, Mark Neumann, Rodney Kinney, and Daniel Weld. 2020.
\newblock \href {https://doi.org/10.18653/v1/2020.acl-main.447} {{S}2{ORC}: The
  semantic scholar open research corpus}.
\newblock In \emph{Proceedings of the 58th Annual Meeting of the Association
  for Computational Linguistics}, pages 4969--4983, Online. Association for
  Computational Linguistics.

\bibitem[{McCann et~al.(2018)McCann, Keskar, Xiong, and
  Socher}]{mccann2018natural}
Bryan McCann, Nitish~Shirish Keskar, Caiming Xiong, and Richard Socher. 2018.
\newblock The natural language decathlon: Multitask learning as question
  answering.
\newblock \emph{arXiv preprint arXiv:1806.08730}.

\bibitem[{McCarthy et~al.(2009)McCarthy, Guess, and
  McNamara}]{mccarthy2009components}
Philip~M McCarthy, Rebekah~H Guess, and Danielle~S McNamara. 2009.
\newblock The components of paraphrase evaluations.
\newblock \emph{Behavior Research Methods}, 41(3):682--690.

\bibitem[{Meuschke(2021)}]{meuschke_norman_2021_4913345}
Norman Meuschke. 2021.
\newblock \href {https://doi.org/10.5281/zenodo.4913345} {\emph{{Analyzing
  Non-Textual Content Elements to Detect Academic Plagiarism}}}.
\newblock Ph.D. thesis, {University of Konstanz, Dept. of Computer and
  Information Science}.
\newblock Doctoral Thesis.

\bibitem[{Mikolov et~al.(2013)Mikolov, Chen, Corrado, and Dean}]{MikolovCCD13}
Tomas Mikolov, Kai Chen, Greg Corrado, and Jeffrey Dean. 2013.
\newblock \href {http://arxiv.org/abs/1301.3781} {Efficient {{Estimation}} of
  {{Word Representations}} in {{Vector Space}}}.
\newblock \emph{ArXiv13013781 Cs}.

\bibitem[{Niu et~al.(2019)Niu, Xu, and Carpuat}]{niu-etal-2019-bi}
Xing Niu, Weijia Xu, and Marine Carpuat. 2019.
\newblock \href {https://doi.org/10.18653/v1/N19-1043} {Bi-directional
  differentiable input reconstruction for low-resource neural machine
  translation}.
\newblock In \emph{Proceedings of the 2019 Conference of the North {A}merican
  Chapter of the Association for Computational Linguistics: Human Language
  Technologies, Volume 1 (Long and Short Papers)}, pages 442--448, Minneapolis,
  Minnesota. Association for Computational Linguistics.

\bibitem[{Pavlick et~al.(2015)Pavlick, Rastogi, Ganitkevitch, Van~Durme, and
  Callison-Burch}]{pavlick-etal-2015-ppdb}
Ellie Pavlick, Pushpendre Rastogi, Juri Ganitkevitch, Benjamin Van~Durme, and
  Chris Callison-Burch. 2015.
\newblock \href {https://doi.org/10.3115/v1/P15-2070} {{PPDB} 2.0: Better
  paraphrase ranking, fine-grained entailment relations, word embeddings, and
  style classification}.
\newblock In \emph{Proceedings of the 53rd Annual Meeting of the Association
  for Computational Linguistics and the 7th International Joint Conference on
  Natural Language Processing (Volume 2: Short Papers)}, pages 425--430,
  Beijing, China. Association for Computational Linguistics.

\bibitem[{Prakash et~al.(2016)Prakash, Hasan, Lee, Datla, Qadir, Liu, and
  Farri}]{PrakashHLD}
Aaditya Prakash, Sadid~A. Hasan, Kathy Lee, Vivek Datla, Ashequl Qadir, Joey
  Liu, and Oladimeji Farri. 2016.
\newblock \href {https://doi.org/10.48550/ARXIV.1610.03098} {Neural paraphrase
  generation with stacked residual lstm networks}.

\bibitem[{Qian et~al.(2019)Qian, Qiu, Zhang, Jiang, and
  Yu}]{qian-etal-2019-exploring}
Lihua Qian, Lin Qiu, Weinan Zhang, Xin Jiang, and Yong Yu. 2019.
\newblock \href {https://doi.org/10.18653/v1/D19-1313} {Exploring diverse
  expressions for paraphrase generation}.
\newblock In \emph{Proceedings of the 2019 Conference on Empirical Methods in
  Natural Language Processing and the 9th International Joint Conference on
  Natural Language Processing (EMNLP-IJCNLP)}, pages 3173--3182, Hong Kong,
  China. Association for Computational Linguistics.

\bibitem[{Raffel et~al.(2020)Raffel, Shazeer, Roberts, Lee, Narang, Matena,
  Zhou, Li, and Liu}]{RaffelSRL20}
Colin Raffel, Noam Shazeer, Adam Roberts, Katherine Lee, Sharan Narang, Michael
  Matena, Yanqi Zhou, Wei Li, and Peter~J. Liu. 2020.
\newblock \href {http://arxiv.org/abs/1910.10683} {Exploring the {{Limits}} of
  {{Transfer Learning}} with a {{Unified Text-to-Text Transformer}}}.
\newblock \emph{ArXiv191010683 Cs Stat}.

\bibitem[{Rush et~al.(2015)Rush, Chopra, and Weston}]{rush-etal-2015-neural}
Alexander~M. Rush, Sumit Chopra, and Jason Weston. 2015.
\newblock \href {https://doi.org/10.18653/v1/D15-1044} {A neural attention
  model for abstractive sentence summarization}.
\newblock In \emph{Proceedings of the 2015 Conference on Empirical Methods in
  Natural Language Processing}, pages 379--389, Lisbon, Portugal. Association
  for Computational Linguistics.

\bibitem[{Shin et~al.(2020)Shin, Razeghi, Logan~IV, Wallace, and
  Singh}]{shin-etal-2020-autoprompt}
Taylor Shin, Yasaman Razeghi, Robert~L. Logan~IV, Eric Wallace, and Sameer
  Singh. 2020.
\newblock \href {https://doi.org/10.18653/v1/2020.emnlp-main.346}
  {{A}uto{P}rompt: {E}liciting {K}nowledge from {L}anguage {M}odels with
  {A}utomatically {G}enerated {P}rompts}.
\newblock In \emph{Proceedings of the 2020 Conference on Empirical Methods in
  Natural Language Processing (EMNLP)}, pages 4222--4235, Online. Association
  for Computational Linguistics.

\bibitem[{Tu et~al.(2017)Tu, Liu, Shang, Liu, and Li}]{AAAI1714161}
Zhaopeng Tu, Yang Liu, Lifeng Shang, Xiaohua Liu, and Hang Li. 2017.
\newblock \href {https://aaai.org/ocs/index.php/AAAI/AAAI17/paper/view/14161}
  {Neural machine translation with reconstruction}.
\newblock \emph{AAAI Conference on Artificial Intelligence}.

\bibitem[{Turing(1950)}]{Turing50}
A.~M. Turing. 1950.
\newblock \href {https://doi.org/10.1093/mind/LIX.236.433}
  {I.\textemdash{{COMPUTING MACHINERY AND INTELLIGENCE}}}.
\newblock \emph{Mind}, LIX(236):433--460.

\bibitem[{Vahtola et~al.(2021)Vahtola, Creutz, Sj{\"o}blom, and
  Itkonen}]{vahtola-etal-2021-coping}
Teemu Vahtola, Mathias Creutz, Eetu Sj{\"o}blom, and Sami Itkonen. 2021.
\newblock \href {https://doi.org/10.18653/v1/2021.wnut-1.32} {Coping with noisy
  training data labels in paraphrase detection}.
\newblock In \emph{Proceedings of the Seventh Workshop on Noisy User-generated
  Text (W-NUT 2021)}, pages 291--296, Online. Association for Computational
  Linguistics.

\bibitem[{Wahle et~al.(2022{\natexlab{a}})Wahle, Ruas, Folt{\'y}nek, Meuschke,
  and Gipp}]{WahleRFM22}
Jan~Philip Wahle, Terry Ruas, Tom{\'a}{\v s} Folt{\'y}nek, Norman Meuschke, and
  Bela Gipp. 2022{\natexlab{a}}.
\newblock \href {https://doi.org/10.1007/978-3-030-96957-8_34} {Identifying
  {{Machine-Paraphrased Plagiarism}}}.
\newblock In Malte Smits, editor, \emph{Information for a {{Better World}}:
  {{Shaping}} the {{Global Future}}}, volume 13192, pages 393--413. {Springer
  International Publishing}, {Cham}.

\bibitem[{Wahle et~al.(2021)Wahle, Ruas, Meuschke, and Gipp}]{Wahle2021}
Jan~Philip Wahle, Terry Ruas, Norman Meuschke, and Bela Gipp. 2021.
\newblock Are {{Neural Language Models Good Plagiarists}}? {{A Benchmark}} for
  {{Neural Paraphrase Detection}}.
\newblock In \emph{Proceedings of the {{ACM}}/{{IEEE Joint Conference}} on
  {{Digital Libraries}} ({{JCDL}})}.

\bibitem[{Wahle et~al.(2022{\natexlab{b}})Wahle, Ruas, Mohammad, and
  Gipp}]{Wahle2022c}
Jan~Philip Wahle, Terry Ruas, Saif~M. Mohammad, and Bela Gipp.
  2022{\natexlab{b}}.
\newblock D3: A massive dataset of scholarly metadata for analyzing the state
  of computer science research.
\newblock In \emph{Proceedings of The 13th Language Resources and Evaluation
  Conference}, Marseille, France. European Language Resources Association.

\bibitem[{Wang and Komatsuzaki(2021)}]{gpt-j}
Ben Wang and Aran Komatsuzaki. 2021.
\newblock {GPT-J-6B: A 6 Billion Parameter Autoregressive Language Model}.
\newblock \url{https://github.com/kingoflolz/mesh-transformer-jax}.

\bibitem[{Witteveen and Andrews(2019)}]{witteveen-andrews-2019-paraphrasing}
Sam Witteveen and Martin Andrews. 2019.
\newblock \href {https://doi.org/10.18653/v1/D19-5623} {Paraphrasing with large
  language models}.
\newblock In \emph{Proceedings of the 3rd Workshop on Neural Generation and
  Translation}, pages 215--220, Hong Kong. Association for Computational
  Linguistics.

\bibitem[{Yuan et~al.(2021)Yuan, Neubig, and Liu}]{yuan2021bartscore}
Weizhe Yuan, Graham Neubig, and Pengfei Liu. 2021.
\newblock Bartscore: Evaluating generated text as text generation.
\newblock \emph{Advances in Neural Information Processing Systems},
  34:27263--27277.

\bibitem[{Zellers et~al.(2019)Zellers, Holtzman, Rashkin, Bisk, Farhadi,
  Roesner, and Choi}]{ZellersHRB19}
Rowan Zellers, Ari Holtzman, Hannah Rashkin, Yonatan Bisk, Ali Farhadi,
  Franziska Roesner, and Yejin Choi. 2019.
\newblock \href {http://arxiv.org/abs/1905.12616} {Defending {{Against Neural
  Fake News}}}.
\newblock \emph{ArXiv190512616 Cs}.

\bibitem[{Zhang et~al.(2019)Zhang, Kishore, Wu, Weinberger, and
  Artzi}]{zhang2019bertscore}
Tianyi Zhang, Varsha Kishore, Felix Wu, Kilian~Q Weinberger, and Yoav Artzi.
  2019.
\newblock Bertscore: Evaluating text generation with bert.
\newblock \emph{arXiv preprint arXiv:1904.09675}.

\bibitem[{Zhou and Bhat(2021)}]{zhou-bhat-2021-paraphrase}
Jianing Zhou and Suma Bhat. 2021.
\newblock \href {https://doi.org/10.18653/v1/2021.emnlp-main.414} {Paraphrase
  generation: A survey of the state of the art}.
\newblock In \emph{Proceedings of the 2021 Conference on Empirical Methods in
  Natural Language Processing}, pages 5075--5086, Online and Punta Cana,
  Dominican Republic. Association for Computational Linguistics.

\end{thebibliography}
